\documentclass[conference]{IEEEtran}
\usepackage{times}
\usepackage{graphicx}
\usepackage{amsmath,amssymb,amsfonts}
\usepackage{array}
\usepackage{textcomp}
\usepackage{url}
\usepackage{cite}
\usepackage{booktabs}
\usepackage{multirow}
\usepackage{xcolor}
\usepackage{caption}
\usepackage[hidelinks]{hyperref}

\hypersetup{
  pdftitle={Aggregation-Induced Calibration Failure in Sum-Based Label Expansion for Multi-Label Facial Expression Recognition: A Diagnostic Study},
  pdfauthor={Weihang Yuan},
  pdfsubject={Multi-label facial expression recognition; label expansion; calibration},
  pdfkeywords={facial expression recognition, multi-label classification, label expansion, calibration, negative results}
}
\definecolor{teal}{HTML}{2F7D6A}
\definecolor{rose}{HTML}{A8563F}

\begin{document}
\title{Aggregation-Induced Calibration Failure in Sum-Based Label Expansion for Multi-Label Facial Expression Recognition: A Diagnostic Study}

\author{\IEEEauthorblockN{Weihang Yuan}\\
\IEEEauthorblockA{Beijing Forestry University, Beijing, China\\
Email: ywh15958108437@bjfu.edu.cn}}

\maketitle

\begin{abstract}
Label-space expansion---predicting a larger joint space and summing back to the original labels---is a natural design for multi-label facial expression recognition, but its failure modes have not been characterised. We evaluate Condition-Aware Label Expansion (CALE) on a self-built dataset (30,240 training images, 3,460 multi-label test images), on seven backbones, and on the public RAF-DB and RAF-ML benchmarks. The central finding is that the aggregation rule, not the representation, decides whether the method works. With a fixed threshold of $0.5$, the aggregate score spans $28\times$ across five backbones at $10\%$ of the training data ($0.24$ to $6.79$) and collapses to a $1.9\times$ range at full data, while threshold-free mAP stays within $1.28\times$ over the same backbones, which localises the fault to the summation step. The mechanism is the uncalibrated rule $p_e = \sigma(x_{e,c^*}) + 0.5\sum_{c\neq c^*}\sigma(x_{e,c})$; we verify it by measuring main- and non-main-branch activations. Two plain seven-dimensional sigmoid baselines (BCE, ASL) reach mAP $0.5055$ and $0.5175$; ASL exceeds three of the five CALE configurations and falls within $0.03$ of the best. On single-label FER the gain over a seven-dimensional baseline is nominally significant (paired $t$-test, $p=0.0436$) but does not survive Benjamini--Hochberg correction ($p_{\text{BH}}=0.2055$) or a Wilcoxon test ($p=0.1250$). We conclude that sum-based label expansion requires threshold calibration at the deployment data scale, or a learned aggregation, before it can support multi-label decisions.
\end{abstract}


\section{Introduction}
	Facial expressions are a primary carrier of emotional and intentional information. The cross-cultural theory of basic emotions by Ekman and Friesen~\cite{ekman1971constants} laid the foundation for automatic FER. Deep convolutional neural networks (CNNs) have achieved significant progress on benchmarks such as FER2013~\cite{goodfellow2013challenges}, RAF-DB~\cite{li2017reliable}, AffectNet~\cite{mollahosseini2019affectnet}, and older datasets such as CK+~\cite{lucey2010extended} and SFEW~\cite{dhall2011static}. Compared with hand-crafted features~\cite{tian2001recognizing}, deep models learn hierarchical representations automatically.
	
	Facial expressions are also a key window into human emotion in psychological research. Psychologists and trained observers typically rely on subjective rating scales, facial action coding systems, or their own experience. Real emotions, however, are rarely pure: a face may simultaneously express ``anger $+$ disgust'' or ``surprise $+$ fear.'' Single-label softmax classification cannot represent such mixtures. Observers' judgments are further affected by image quality, illumination, and resolution, making consistent annotation difficult. An automatic system that handles mixed emotions and degraded images would therefore be a valuable tool for psychologists, providing objective and reproducible assessments.
	
	Real-world FER faces four challenges: (1) \textbf{image quality degradation}---uneven illumination, low resolution, and blur can make the same expression look more different than two distinct expressions~\cite{li2017reliable}; (2) \textbf{coexisting multiple emotions}---traditional single-label classification cannot represent mixtures; (3) \textbf{imbalanced class distributions}---rare expressions such as disgust and fear suffer from low recall~\cite{zhang2021relative}; and (4) \textbf{training-inference inconsistency}---multi-branch models that train only the current condition branch but sum all branches at inference face a mismatched objective~\cite{wang2020suppressing}.
	
	To address these issues, we propose and systematically study Condition-Aware Label Expansion (CALE), a label-space organization strategy that encodes image quality conditions into the label space and uses a multi-branch joint output with summation aggregation. Rather than claiming a new state-of-the-art method, this paper provides a diagnostic empirical study of when and why this strategy helps. Our main contribution is the diagnostic analysis: the mechanism itself is described for completeness, but the central claims of the paper are the empirical findings and their practical implications. The main contributions are summarized as follows:
	
	\begin{itemize}
		\item \textbf{A diagnostic study of sum-based label expansion.} We present a systematic evaluation of CALE, a concrete instantiation of the general design pattern $p_e = \sum_c w_c \sigma(x_{e,c})$ with $w_{c^*} = 1$ and $w_{c \neq c^*} = 0.5$. We frame CALE as a case study, not as a state-of-the-art method.
		\item \textbf{An aggregation-induced calibration failure.} Across four training-data regimes ($10\%$, $25\%$, $50\%$, $100\%$) and five backbones, we show that the aggregate score at threshold $0.5$ varies by $28\times$ across backbones at $10\%$ data and collapses to systematic under-prediction at full data. We verify the mechanism by direct measurement of the main-branch and non-main-branch sigmoid activations, and we show that the threshold-free mAP is stable across the same backbones, which localizes the failure to the aggregation step.
		\item \textbf{A fair comparison against non-condition-aware baselines.} Two seven-dimensional sigmoid baselines (BCE and ASL) reach mAP $0.5055$ and $0.5175$ under the same protocol; the ASL baseline exceeds three of the five CALE configurations and falls within $0.03$ of the best. This is direct evidence that the additional condition dimensions do not translate into a consistent architecture-independent advantage on multi-label FER.
		\item \textbf{A quantified limitation of the rule-based condition classifier.} A $\pm 20\%$ threshold perturbation shifts the dominant condition class by nearly $29$ percentage points, which is a first-order sensitivity for a component that determines both the training soft-label target and the inference branch weights.
		\item \textbf{Honest reporting of three structural limitations.} We document a single-label training / multi-label evaluation mismatch, a coupling between the synthetic-condition generator and the rule-based condition classifier, and single-fold single-seed diagnostics. We do not claim that these limitations are resolved in the present study.
		\item \textbf{Full reproducibility.} Code, data-split indices, trained model weights, and a full description of the diagnostic experiments are publicly available.
	\end{itemize}
	
	The remainder of this paper is organized as follows. Section~II reviews related work. Section~III presents the CALE method. Section~IV reports the main experiments. Section~V presents the diagnostic analysis. Section~VI discusses findings and applications, and Section~VII concludes the paper.
	
	\section{Related Work}
	
	\subsection{Deep Facial Expression Recognition}
	Mollahosseini et al.~\cite{mollahosseini2016going} first applied deep CNNs to seven-class FER, and Rifai et al.~\cite{rifai2012disentangling} explored disentangled factor representations for the same task. He et al.~\cite{he2016deep} proposed ResNet, which has become a standard backbone. Hu et al.~\cite{hu2018squeeze} proposed SE networks. Recent works have explored vision transformers~\cite{vaswani2017attention,dosovitskiy2021image}. To handle uncertainty, Wang et al.~\cite{wang2020suppressing} proposed SCN; Ruan et al.~\cite{ruan2021feature} proposed feature decomposition; Zhao et al.~\cite{zhao2021learning} used multi-scale attention; Liu et al.~\cite{liu2022adaptive} proposed adaptive multilayer attention; Xue et al.~\cite{xue2021transfer} introduced TransFER, and Wang et al.~\cite{wang2020region} proposed region attention networks for pose and occlusion robustness. These methods, however, assume stable image quality and pay limited attention to condition robustness.
	
	\subsection{Multi-Label Classification}
	Multi-label learning allows a sample to belong to multiple classes~\cite{zhang2014review,tsoumakas2010mining}. Deep methods typically use sigmoid with binary cross-entropy. Recent work has focused on loss design, including Asymmetric Loss~\cite{ridnik2021asymmetric} and Class-Balanced Loss~\cite{cui2019class}; graph-based methods such as ML-GCN~\cite{chen2019multi} model label co-occurrence explicitly. In FER, multi-label learning is still emerging. Li and Deng~\cite{li2019blended} introduced RAF-ML for blended emotion recognition. We embed condition information into the label space via label expansion, so the multi-branch structure naturally supports multi-label output.
	
	\subsection{Quality-Aware Learning}
	Quality-aware learning improves robustness to low-quality inputs via enhancement, multi-branch networks, quality estimation, or domain adaptation. Li et al.~\cite{li2017reliable} showed that image quality significantly affects recognition, and classical image-quality metrics such as SSIM~\cite{wang2004image} remain widely used to quantify degradation. Unlike these works, we encode quality directly into the label space without extra estimation or reconstruction.
	
	\subsection{Attention Mechanisms}
	Attention focuses on key information~\cite{bahdanau2015neural}. SE obtains channel descriptors via global average pooling and learns channel weights~\cite{hu2018squeeze}. CBAM adds spatial attention~\cite{woo2018cbam}. Non-local networks model long-range dependencies~\cite{wang2018non}. We adopt SE for its simplicity, few parameters, and training stability.
	
	\section{Method}
	
	\subsection{Dataset Construction}
	The self-built dataset is constructed from publicly available images and video frames. Specifically, the source material comes from two types of public content: (1) films and TV series, and (2) short videos posted by online content creators. All images contain only publicly accessible human faces that have been voluntarily shown in public media. No private or sensitive personal information is involved, and the images are used solely for non-commercial academic research. This usage follows the same principle adopted by public FER datasets such as RAF-DB and AffectNet.
	
	\begin{table}[t]
		\centering
		\caption{Training Set Sample Distribution}
		\label{tab:train_dist}
		\begin{tabular}{lc}
			\toprule
			\textbf{Emotion} & \textbf{Number of Samples} \\
			\midrule
			Angry     & 4,320 \\
			Disgust   & 4,320 \\
			Fear      & 4,320 \\
			Happy     & 4,320 \\
			Sad       & 4,320 \\
			Surprise  & 4,320 \\
			Natural   & 4,320 \\
			\midrule
			\textbf{Total} & \textbf{30,240} \\
			\bottomrule
		\end{tabular}
	\end{table}
	
	For each of the seven basic emotions (angry, disgust, fear, happy, sad, surprise, natural), 216 source clips or stills are selected, forming a baseline set of $7 \times 216 = 1{,}512$ images. Each source image is treated as one standard-condition sample, and 19 additional variants are generated through five geometric and photometric transformations (rotation, translation, brightness adjustment, contrast adjustment, and Gaussian blur) across four quality conditions. Each source image therefore yields 20 samples, and each class contains $216 \times 20 = 4{,}320$ samples. The seven classes together contain 30{,}240 training images. All images are aligned by MTCNN~\cite{zhang2016joint} and resized to $224 \times 224$.
	
	The test set consists of 3{,}460 independently collected multi-label images from different sources, and each test image may belong to more than one emotion category. Table~\ref{tab:train_dist} lists the training distribution, and Table~\ref{tab:cond_dist} reports the condition distribution. We treat each raw photo and its 19 derived variants as one group; 5-fold cross-validation is grouped by this identifier to prevent leakage.
	
	\begin{table}[t]
		\centering
		\caption{Condition Distribution of Training and Test Sets}
		\label{tab:cond_dist}
		\begin{tabular}{lcc}
			\toprule
			\textbf{Condition} & \textbf{Training Set} & \textbf{Test Set} \\
			\midrule
			Weak light   & 7,520 & 949 \\
			Strong light & 9,565 & 1,003 \\
			Low resolution & 5,897 & 678 \\
			Standard     & 7,258 & 830 \\
			\midrule
			\textbf{Total} & \textbf{30,240} & \textbf{3,460} \\
			\bottomrule
		\end{tabular}
	\end{table}
	
	\subsection{Overall Framework}
\begin{figure*}[t]
		\centering
		\includegraphics[width=0.95\textwidth]{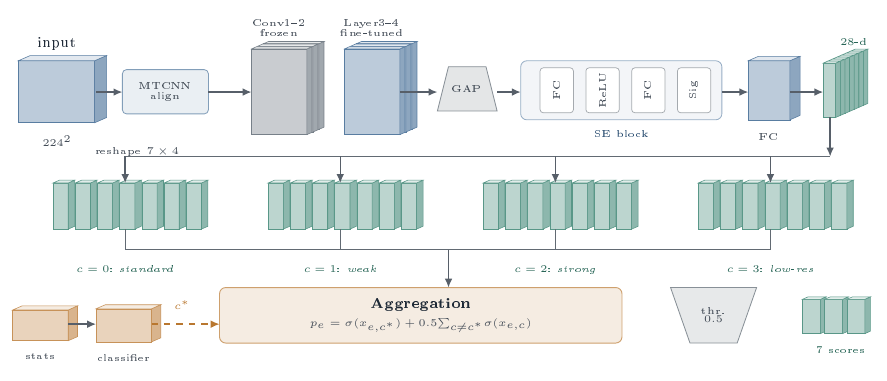}
		\caption{Overall framework of the CALE method. After MTCNN alignment, the image is processed by a shared backbone with shallow layers frozen. A global average pooling layer and an SE module recalibrate channel features, followed by flatten and dropout. The 28-dimensional output is organized as four condition-specific 7-dimensional branches. Each branch is sigmoid-activated, and the four branches are aggregated per emotion to produce seven scores. A fixed threshold yields the final multi-label prediction. This figure was drawn by the author with TikZ.}
		\label{fig:framework}
	\end{figure*}

	Fig.~\ref{fig:framework} illustrates the overall framework. Each raw photo is augmented across four quality conditions, producing 19 variants plus the original. The condition classifier assigns one of the four quality conditions to each image using the fixed-threshold rule in Table~\ref{tab:cond_thresh}. This label is used in two places: during training, it determines which of the four condition positions is treated as the main condition $c^*$ in Eq.~(4); during inference, it determines the branch weight in Eq.~(5). Because the classifier is rule-based and deterministic, the same image receives the same label at training and inference time, which preserves the training-inference consistency that is central to CALE. We characterize its behavior in Section~\ref{sec:cond_classifier} and show that its output is extremely sensitive to the thresholds, which we identify as one of the method's limitations. After MTCNN~\cite{zhang2016joint} alignment, images pass through a shared backbone in which shallow layers are frozen and only the deeper layers are fine-tuned. The output then passes through global average pooling and an SE module, followed by dropout and a fully connected layer to produce a 28-dimensional vector. Each group of four dimensions corresponds to one emotion under four conditions. At inference, the 28 logits are aggregated per emotion into seven scores, and a fixed threshold of $0.5$ yields the final multi-label prediction.
	
	\subsection{Condition Classifier}
	The classifier assigns each image to one of four conditions using two inexpensive statistics:
	
	\begin{itemize}
		\item \textbf{Brightness mean}: average pixel value, reflecting illumination.
		\item \textbf{Laplacian variance}: variance of the Laplacian response, measuring sharpness~\cite{pech2000diatom}.
	\end{itemize}
	
	The rules are listed in Table~\ref{tab:cond_thresh}. Thresholds are set empirically, and condition labels are precomputed once per image. The classifier is used for two purposes: (i) in training, it determines the main condition position $c^*$ in the soft-label target in Eq.~(4); (ii) in inference, it supplies the one-hot vector that weights the four branches in Eq.~(5). We report the classifier's behavior and its threshold sensitivity in Section~\ref{sec:cond_classifier}.
	
	\begin{table}[t]
		\centering
		\caption{Condition Classification Thresholds}
		\label{tab:cond_thresh}
		\begin{tabular}{ll}
			\toprule
			\textbf{Condition} & \textbf{Rule} \\
			\midrule
			Weak light   & Lap.\ variance $\geq 10$ and brightness mean $< 80$ \\
			Strong light & Lap.\ variance $\geq 10$ and brightness mean $> 120$ \\
			Low resolution & Lap.\ variance $< 10$ \\
			Standard     & otherwise \\
			\bottomrule
		\end{tabular}
	\end{table}
	
	\subsection{Feature Extraction and SE Attention}
	We adopt ImageNet-pretrained backbones. Shallow layers are frozen; only the deeper layers, the SE module, and the fully connected layer are fine-tuned. This preserves generic features and reduces the number of trainable parameters, lowering the risk of overfitting.
	
	After the backbone outputs a feature map, the SE module performs global average pooling:
	\begin{equation}
		z_{c} = \frac{1}{H\times W}\sum_{i = 1}^{H}\sum_{j = 1}^{W}F_{c}(i,j).
	\end{equation}
	
	Channel dependencies are then learned as:
	\begin{equation}
		\mathbf{s} = \sigma \left(\mathbf{W}_2\delta (\mathbf{W}_1\mathbf{z})\right),
	\end{equation}
	where $\mathbf{W}_1 \in \mathbb{R}^{\frac{C}{r}} \times C$, $\mathbf{W}_2 \in \mathbb{R}^{C \times \frac{C}{r}}$, $r = 16$, $\delta$ is ReLU, and $\sigma$ is Sigmoid. The recalibration is:
	\begin{equation}
		\mathbf{F}_{se} = \mathbf{F}\odot \mathbf{s},
	\end{equation}
	where $\odot$ denotes channel-wise multiplication. The SE module enhances channels beneficial for expression recognition and suppresses irrelevant ones.
	
\begin{figure}[t]
		\centering
		\includegraphics[width=\columnwidth]{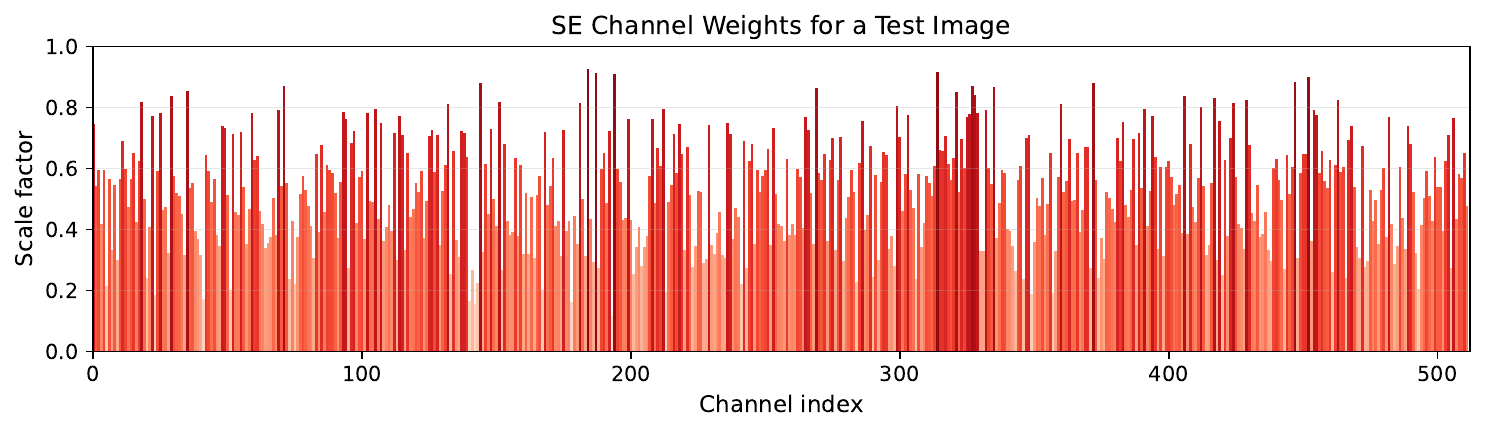}
		\caption{Visualization of the 512-dimensional SE scale factors for a test image. Higher values indicate channels more strongly enhanced by the SE module.}
		\label{fig:se}
	\end{figure}

	To verify the SE module, we visualize the 512-dimensional scale factors for a test image (Fig.~\ref{fig:se}). Some channels have factors close to 1, while many others are close to 0, indicating selective enhancement rather than uniform weighting.
	
	\subsection{Joint Output and Loss Function}
	After SE, features pass through dropout (rate $0.5$) and a linear layer to produce a 28-dimensional output ($7$ emotions $\times$ $4$ conditions). We refer to this condition-aware label expansion as LE. The logit for emotion $e$ under condition $c$ is $x_{e,c}$, and the sigmoid-activated probability is $\sigma (x_{e,c})$.
	
	During training, the main condition position of the ground-truth emotion is set to $1.0$, and the other condition positions are set to $0.5$:
	\begin{equation}
		y_{e,c} =
		\begin{cases}
			1.0, & c = c^*, e = e_{gt} \\
			0.5, & c \neq c^*, e = e_{gt} \\
			0,   & e \neq e_{gt}
		\end{cases}
	\end{equation}
	
	During inference, the four condition scores are aggregated by summation:
	\begin{equation}
		p_{e} = \sigma (x_{e,c^*}) + 0.5 \sum_{c \neq c^*}\sigma (x_{e,c}),
	\end{equation}
	where $p_{e} \in [0, 2.5]$ is not normalized to a probability. A fixed threshold of $0.5$ produces the final multi-label prediction. This aggregation structurally matches the soft-label training.
	
	For multi-strategy training, we use a soft-label variant of Focal Loss:
	\begin{equation}
		\mathcal{L} = - \alpha_t (1-p_t)^\gamma \log(p_t),
	\end{equation}
	where $p_t$ is the model's estimated probability for the target class, $\gamma=2.0$ is the focusing parameter, and $\alpha_t$ is the class-weighted coefficient. Weighted sampling, strong augmentation, dropout $0.5$, and shallow-layer freezing form the remaining components of MS.
	\section{Experiments}
	
	\subsection{Datasets and Protocol}
	We evaluate CALE on three datasets:
	\begin{itemize}
		\item \textbf{Self-built dataset}: 30{,}240 training images (7 emotions, 4 conditions) and 3{,}460 independently collected multi-label test images.
		\item \textbf{RAF-DB}~\cite{li2017reliable}: 12{,}271 training and 3{,}068 test images with 7 basic emotions.
		\item \textbf{RAF-ML}~\cite{li2019blended}: 3{,}926 training and 982 test images with blended multi-label annotations across 6 emotions.
	\end{itemize}
	
	For 5-fold cross-validation on the self-built dataset, subjects are grouped so that variants of the same source image never appear in both training and validation. For cross-backbone experiments, we split the training set into $7/8$ training and $1/8$ validation, with 27 subjects per emotion in the validation set.
	
	\subsection{Implementation Details}
	All models are trained with AdamW using a learning rate of $1\times 10^{-5}$, a weight decay of $5\times 10^{-3}$, and cosine annealing. The batch size is $128$ for ResNet-18/34 and $96$ for ResNet-50/DenseNet121. Focal Loss uses $\gamma=2.0$ and class weights $[0.9, 0.8, 1.0, 0.3, 0.5, 0.7, 0.2]$ for the seven emotions. The random seed is fixed at $42$ across all experiments. All experiments were performed on a workstation running Ubuntu 20.04 with an NVIDIA RTX 3090 GPU (24GB), using PyTorch 1.10 and Python 3.8.
	
	\subsection{Main Results on the Self-Built Dataset}
	Table~\ref{tab:main} reports the 5-fold single-label cross-validation results. CALE (V7) achieves a Macro-F1 of $0.7747$, compared with $0.7618$ for the V0 baseline ($+1.29$ points). The best single-label variant is V5 (LE+MS) at $0.7758$.
	
	\begin{table*}[t]
		\centering
		\caption{Main results on the self-built dataset (5-fold single-label cross-validation, mean $\pm$ std). LE: label expansion; MS: multi-strategy training; SE: squeeze-and-excitation.}
		\label{tab:main}
		\begin{tabular}{lcccccc}
			\toprule
			\textbf{Variant} & \textbf{LE} & \textbf{MS} & \textbf{SE} & \textbf{Val Accuracy} & \textbf{Val UAR} & \textbf{Val Macro-F1} \\
			\midrule
			V0: Baseline     & -- & -- & -- & $0.7635 \pm 0.0107$ & $0.7634 \pm 0.0106$ & $0.7618 \pm 0.0118$ \\
			V1: +MS          & -- & \checkmark & -- & $0.7716 \pm 0.0102$ & $0.7715 \pm 0.0098$ & $0.7705 \pm 0.0098$ \\
			V2: +LE          & \checkmark & -- & -- & $0.7745 \pm 0.0078$ & $0.7745 \pm 0.0077$ & $0.7729 \pm 0.0082$ \\
			V3: +SE          & -- & -- & \checkmark & $0.7726 \pm 0.0086$ & $0.7725 \pm 0.0084$ & $0.7706 \pm 0.0085$ \\
			V4: +MS+SE       & -- & \checkmark & \checkmark & $0.7723 \pm 0.0093$ & $0.7722 \pm 0.0090$ & $0.7709 \pm 0.0088$ \\
			V5: +LE+MS       & \checkmark & \checkmark & -- & $\mathbf{0.7773 \pm 0.0131}$ & $\mathbf{0.7772 \pm 0.0129}$ & $\mathbf{0.7758 \pm 0.0128}$ \\
			V6: +LE+SE       & \checkmark & -- & \checkmark & $0.7764 \pm 0.0107$ & $0.7763 \pm 0.0104$ & $0.7750 \pm 0.0105$ \\
			\textbf{V7: Full} & \checkmark & \checkmark & \checkmark & $0.7764 \pm 0.0149$ & $0.7763 \pm 0.0147$ & $0.7747 \pm 0.0148$ \\
			\bottomrule
		\end{tabular}
	\end{table*}

	\subsection{Ablation Study}
	We conduct a $2^3$ factorial ablation study with ResNet-18. For brevity, we denote label expansion as LE, multi-strategy training as MS, and Squeeze-and-Excitation attention as SE. Table~\ref{tab:main} reports the full comparison, and Table~\ref{tab:ttest} reports paired tests across the 5 folds.
	
	\textbf{Label expansion is the primary driver.} All variants containing LE (V2, V5, V6, and V7) achieve Val Macro-F1 $\geq 0.7747$, while variants without LE (V0, V1, V3, and V4) stay below $0.7729$. LE alone (V2 vs V0) improves Validation UAR by $+0.0078$ ($t=2.917$, $p=0.0436$ under an uncorrected paired $t$-test).
	
	\textbf{SE provides a robustness benefit.} V6 (LE+SE vs V0) achieves the strongest paired mean improvement ($+0.0096$, $t=3.270$, $p=0.0308$). Compared to V2 alone, V6 adds only $+0.0018$ in the mean but reduces the fold-to-fold variance, making the improvement over the baseline more consistent.
	
	\textbf{MS increases variance without a statistically separable gain.} Although V5 (LE+MS) achieves the highest mean Validation UAR at $0.7758$, its paired improvement over V0 ($+0.0105$) does not reach significance ($p=0.0589$), and the full model V7 is even less significant ($p=0.1672$) because its fold-to-fold variance is the largest among all variants ($\pm 0.0148$ in Val Macro-F1). Pairwise comparisons among LE-containing variants (V5 vs V2, V6 vs V2, V7 vs V2) are all non-significant ($p > 0.5$). This suggests that MS is associated with a higher mean but also with greater fold-to-fold variability, so its net effect is not statistically separable from LE alone with 5 folds.
	
	\textbf{Non-parametric robustness check and multiple-comparison correction.} Because paired $t$-tests assume normality and $n=5$, and because the table reports $5$ computable comparisons, we apply two robustness checks. First, under the Wilcoxon signed-rank test, none of the comparisons against V0 reach significance: V2 vs V0 gives $p=0.125$, V3 vs V0 gives $p=0.188$, V4 vs V0 gives $p=0.188$, and V7 vs V0 gives $p=0.188$. Second, under a Benjamini--Hochberg FDR correction at $q=0.05$ over the $5$ computable comparisons, the paired $t$-test adjusted $p$-values are V2 vs V0 $p_{\text{BH}}=0.2055$, V3 vs V0 $p_{\text{BH}}=0.2055$, V4 vs V0 $p_{\text{BH}}=0.2055$, and V7 vs V0 $p_{\text{BH}}=0.2090$, none of which is significant. The nominally significant improvement of LE therefore does not survive either a non-parametric test or a multiple-comparison correction, and the single-label advantage of CALE should be treated as exploratory rather than established.
	
	\textbf{Multi-label performance at target density.} We evaluate on the multi-label test set with threshold scanning to match the true label density ($2.5$). Table~\ref{tab:multilabel_ablation} reports the results. MS alone (V0 $\to$ V1) is associated with a $+0.0316$ change in multi-label Macro-F1, while LE alone is associated with $+0.0154$. Adding MS on top of LE (V2 $\to$ V5) is associated with $+0.0178$.
	
	\begin{table}[t]
		\centering
		\caption{Multi-label test set results at $\text{AvgLabels} \approx 2.5$.}
		\label{tab:multilabel_ablation}
		\begin{tabular}{lcccc}
			\toprule
			\textbf{Variant} & \textbf{LE} & \textbf{MS} & \textbf{SE} & \textbf{Macro-F1} \\
			\midrule
			V0 & -- & -- & -- & 0.5607 \\
			V1 & -- & \checkmark & -- & 0.5923 \\
			V2 & \checkmark & -- & -- & 0.5761 \\
			V3 & -- & -- & \checkmark & 0.5643 \\
			V4 & -- & \checkmark & \checkmark & 0.5928 \\
			V5 & \checkmark & \checkmark & -- & 0.5939 \\
			V6 & \checkmark & -- & \checkmark & 0.5844 \\
			\textbf{V7} & \checkmark & \checkmark & \checkmark & \textbf{0.5966} \\
			\bottomrule
		\end{tabular}
	\end{table}
	
	\begin{table}[t]
		\centering
		\caption{Paired tests across the 5 folds on Validation UAR. V0 is a reproduction with Val UAR $=0.7667$. Columns ``$t$/$p_t$'' are a paired $t$-test; column ``$p_W$'' is a Wilcoxon signed-rank test. After Benjamini--Hochberg FDR correction at $q=0.05$ over the $5$ computable comparisons, none of the paired $t$-test comparisons are significant ($p_{\text{BH}} \geq 0.2055$); the same holds for the Wilcoxon comparisons ($p_{\text{BH}} \geq 0.2344$). We therefore do not claim a statistically established single-label advantage.}
		\label{tab:ttest}
		\footnotesize
		\setlength{\tabcolsep}{3pt}
		\begin{tabular}{lccccc}
			\toprule
			\textbf{Comparison} & \textbf{$\Delta$ mean} & \textbf{$t$} & \textbf{$p_t$} & \textbf{$p_W$} \\
			\midrule
			V2 (LE) vs V0        & $+0.0078$ & $2.917$ & $0.0436$ & $0.1250$ \\
			V3 (SE) vs V0        & $+0.0058$ & $1.947$ & $0.1233$ & $0.1875$ \\
			V4 (MS+SE) vs V0     & $+0.0055$ & $2.046$ & $0.1102$ & $0.1875$ \\
			V5 (LE+MS) vs V0     & $+0.0105$ & $2.614$ & $0.0589$ & --- \\
			V6 (LE+SE) vs V0     & $+0.0096$ & $3.270$ & $0.0308$ & --- \\
			V7 (Full) vs V0      & $+0.0096$ & $1.685$ & $0.1672$ & $0.1875$ \\
			V5 vs V2             & $+0.0027$ & $0.663$ & $0.5436$ & --- \\
			V6 vs V2             & $+0.0018$ & $0.479$ & $0.6569$ & --- \\
			V7 vs V2             & $+0.0018$ & $0.238$ & $0.8236$ & $1.0000$ \\
			V7 vs V5             & $-0.0009$ & $-0.149$ & $0.8893$ & --- \\
			\bottomrule
		\end{tabular}
	\end{table}

	\subsection{Cross-Backbone Evaluation}
	We evaluate CALE on seven mainstream backbones. Table~\ref{tab:crossbackbone} reports the single-label and multi-label results on the self-built dataset ($1/8$ validation split).
	
	\begin{table*}[t]
		\centering
		\caption{Cross-backbone evaluation. Single-label results use the $1/8$ validation split; multi-label results use the independent test set at $\text{AvgLabels} \approx 2.5$. The multi-label column uses a threshold scan that matches the target label density on the evaluation set; these numbers should be interpreted as an upper bound rather than as achievable performance (see Section~\ref{sec:diag_scope}).}
		\label{tab:crossbackbone}
		\begin{tabular}{lcccccc}
			\toprule
			\multirow{2}{*}{Backbone} & \multirow{2}{*}{Params (M)} & \multicolumn{2}{c}{Single-label Macro-F1} & \multicolumn{2}{c}{Multi-label Macro-F1} \\
			\cmidrule(lr){3-4} \cmidrule(lr){5-6}
			& & \textbf{Baseline} & \textbf{+CALE} & \textbf{Baseline} & \textbf{+CALE} \\
			\midrule
			ResNet-18       & 11.2 & 0.7830 & 0.7951 & 0.5596 & 0.5959 \\
			ResNet-34       & 21.8 & 0.7829 & 0.8016 & 0.5637 & 0.5965 \\
			ResNet-50       & 25.6 & 0.7834 & 0.8011 & 0.5566 & 0.5956 \\
			MobileNetV2     & 3.5  & 0.7726 & 0.7859 & 0.5321 & 0.5948 \\
			DenseNet121     & 8.0  & 0.7756 & 0.7977 & 0.5379 & 0.5985 \\
			EfficientNet-B0 & 5.3  & 0.8029 & \textbf{0.8099} & 0.5667 & \textbf{0.6023} \\
			DeiT-Tiny       & 5.7  & 0.7749 & 0.7631 & 0.5157 & 0.5745 \\
			\bottomrule
		\end{tabular}
	\end{table*}
	
	Three observations follow. First, CALE improves single-label performance on 6 of 7 backbones by $+0.70$ to $+2.21$ points, with EfficientNet-B0 + CALE achieving the best single-label result of $0.8099$. Second, under the full-data multi-condition setting with threshold scanning to match the target label density, CALE is associated with multi-label improvements of $+3.28$ to $+6.27$ points. Third, DeiT-Tiny is the only exception on single-label performance ($-1.18$ points). The cross-backbone picture under the threshold-scanned setting is favorable; we show below that this picture depends on threshold calibration, and that a fixed threshold produces very different behavior under different training data scales.

	\subsection{Public Benchmarks}
	
	\subsubsection{RAF-DB}
	We evaluate CALE on RAF-DB. Table~\ref{tab:rafdb} reports the results. With raw RAF-DB training data (single condition), CALE drops to a Macro-F1 of $0.5226$. After augmenting the training set with four-condition variants (49{,}084 images), CALE recovers to $0.7475$, exceeding the raw baseline ($0.6769$) by $+7.1$ points but remaining below the multi-condition baseline ($0.7603$).
	
	\begin{table}[t]
		\centering
		\caption{RAF-DB evaluation ($1/8$ split used for baseline; augmented training for multi-condition settings). The row ``V1 diagnostic (raw)'' corresponds to a 7-dimensional baseline trained with the multi-strategy training (MS) configuration on raw RAF-DB.}
		\label{tab:rafdb}
		\begin{tabular}{lcc}
			\toprule
			\textbf{Setting} & \textbf{Training Data} & \textbf{Test Macro-F1} \\
			\midrule
			Baseline (raw)               & 12K & 0.6769 \\
			V1 diagnostic (raw)          & 12K & 0.7555 \\
			CALE (raw)                   & 12K & 0.5226 \\
			Baseline (multi-condition)   & 49K & \textbf{0.7603} \\
			CALE (multi-condition)       & 49K & 0.7475 \\
			\bottomrule
		\end{tabular}
	\end{table}
	
	\subsubsection{RAF-ML}
	Table~\ref{tab:rafml} reports the multi-label benchmark. With raw training data, CALE achieves a Macro-F1 of $0.6654$ at the calibrated threshold ($0.99$), below our own baseline ($0.6835$). With multi-condition training, CALE reaches $0.7330$ at the fixed threshold $0.5$ and $0.7207$ at $\text{AvgLabels} = 2.20$, both below the multi-condition baseline ($0.7359$ at $\text{AvgLabels} = 2.08$).
	
	\begin{table}[t]
		\centering
		\caption{RAF-ML multi-label evaluation under the fixed decision threshold $0.5$. All rows use the fixed decision threshold $0.5$; threshold-calibrated variants are not reported because they require evaluation-set label density.}
		\label{tab:rafml}
		\footnotesize
		\setlength{\tabcolsep}{4pt}
		\begin{tabular}{lcc}
			\toprule
			\textbf{Setting} & \textbf{AvgLabels} & \textbf{Macro-F1} \\
			\midrule
			Prior BCE baseline~\cite{li2019blended}   & --     & 0.5954 \\
			Prior ASL~\cite{ridnik2021asymmetric}     & --     & 0.6481 \\
			Prior ASL with soft labels~\cite{ridnik2021asymmetric} & -- & 0.6500 \\
			\midrule
			Baseline (raw)                 & 2.06 & 0.6835 \\
			CALE (raw, Focal)              & 5.61 & 0.5526 \\
			Baseline (multi-cond.)         & 2.08 & \textbf{0.7359} \\
			CALE (multi-cond., Focal)      & 4.36 & 0.6411 \\
			CALE (multi-cond., BCE)        & 2.45 & 0.7330 \\
			\bottomrule
		\end{tabular}
	\end{table}
	
	\subsubsection{Threshold Sensitivity}
\begin{figure*}[t]
		\centering
		\includegraphics[width=0.95\textwidth]{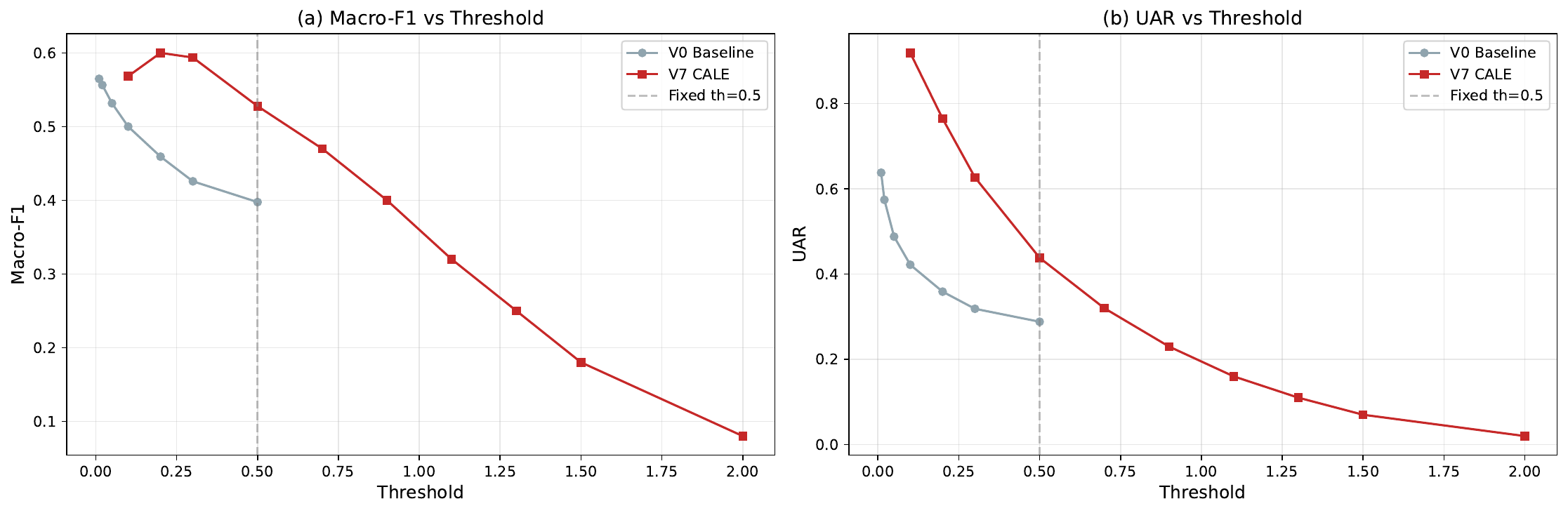}
		\caption{Threshold sensitivity on the multi-label test set. (a) Macro-F1 vs threshold. (b) UAR vs threshold. The CALE curve spans a wider threshold range because the aggregation score is not normalized, not because the model predicts more labels correctly.}
		\label{fig:threshold}
	\end{figure*}

	Fig.~\ref{fig:threshold} plots Macro-F1 and UAR as functions of the decision threshold for the baseline (V0) and the full CALE model (V7). The baseline softmax curve saturates at $\text{AvgLabels} = 1.0$ once the threshold exceeds $0.5$. In contrast, the CALE aggregation score spans $[0, 2.5]$, and the curve decreases smoothly from $\text{AvgLabels} = 5.6$ at $0.5$ to $1.0$ at $1.2$. This behavior reflects the range of the aggregated score rather than a performance advantage, as we analyze in Section~\ref{sec:diagnostic}.
	
	\subsection{Condition-wise Analysis}
	Table~\ref{tab:condwise} reports per-condition UAR on the augmented RAF-DB test set. CALE shows comparable performance across all four conditions, with weak light ($0.7482$) slightly exceeding standard ($0.7437$).
	
	\begin{table}[t]
		\centering
		\caption{Per-condition UAR on augmented RAF-DB test set.}
		\label{tab:condwise}
		\begin{tabular}{lcc}
			\toprule
			\textbf{Condition} & \textbf{Baseline} & \textbf{CALE} \\
			\midrule
			Standard       & 0.7555 & 0.7437 \\
			Weak light     & 0.7514 & \textbf{0.7482} \\
			Strong light   & 0.7078 & 0.6925 \\
			Low resolution & 0.7142 & 0.7190 \\
			\bottomrule
		\end{tabular}
	\end{table}

	\subsubsection{Per-emotion Analysis}
	Table~\ref{tab:peremotion} reports per-emotion F1 on the RAF-DB test set. CALE improves five of the seven categories, most on \textit{happy} ($+0.0519$) and \textit{disgust} ($+0.0251$), and loses slightly on \textit{angry} ($-0.0165$) and \textit{fear} ($-0.0237$). The mean gain is $+0.0121$, consistent with the $+1.29$ points reported for the full model.
	
	\begin{table}[t]
	\centering
	\caption{Per-emotion F1 on the RAF-DB test set ($1/8$ split). $\Delta$ is CALE minus baseline.}
	\label{tab:peremotion}
	\footnotesize
	\setlength{\tabcolsep}{4pt}
	\begin{tabular}{lccc}
		\toprule
		\textbf{Emotion} & \textbf{Baseline} & \textbf{CALE} & \textbf{$\Delta$} \\
		\midrule
		Angry    & 0.8070 & 0.7905 & \textcolor{rose}{$-0.0165$} \\
		Disgust  & 0.7044 & \textbf{0.7295} & \textcolor{teal}{$+0.0251$} \\
		Fear     & 0.8247 & 0.8010 & \textcolor{rose}{$-0.0237$} \\
		Happy    & 0.8458 & \textbf{0.8977} & \textcolor{teal}{$+0.0519$} \\
		Sad      & 0.7817 & 0.7925 & \textcolor{teal}{$+0.0108$} \\
		Surprise & 0.7062 & 0.7233 & \textcolor{teal}{$+0.0171$} \\
		Natural  & 0.8111 & 0.8311 & \textcolor{teal}{$+0.0200$} \\
		\midrule
		Mean     & 0.7830 & 0.7951 & \textcolor{teal}{$+0.0121$} \\
		\bottomrule
	\end{tabular}
\end{table}
	
	\section{Diagnostic Analysis: When Does CALE Help?}
	\label{sec:diagnostic}
	
	The results in Sections~IV-C to IV-F show a mixed pattern: CALE improves single-label FER, but its multi-label behavior is not uniform. To understand why, we conduct a set of diagnostic experiments under four training-data regimes ($10\%$, $25\%$, $50\%$, $100\%$ of the training set), a single fold, and five backbones (ResNet-18/34/50, MobileNetV2, EfficientNet-B0). The reduced-data regimes amplify data-scale effects and keep the diagnostic cost manageable.
	
	\subsection{Scope and Limitations}
	\label{sec:diag_scope}
	We emphasize that the diagnostics are preliminary and cover a single fold, four data regimes, and a limited set of backbones. The behavior reported below may differ at additional backbones. We frame the findings as directional evidence for a working hypothesis, not as a proven causal mechanism, and identify the additional experiments needed to test the hypothesis in Section~\ref{sec:future_work}.
	
	We additionally note three limitations of the diagnostic study that are important for interpreting the results. First, the multi-label evaluation uses an uncalibrated threshold of $0.5$; a threshold tuned on a held-out multi-label validation set would change the absolute AvgLabels and may change some of the qualitative conclusions. The threshold scans used in Tables~\ref{tab:multilabel_ablation}, \ref{tab:crossbackbone}, and \ref{tab:rafml} match the target label density on the evaluation set and therefore use information that would not be available at deployment time; those numbers should be interpreted as upper bounds rather than as achievable performance. Second, the diagnostics use a single random seed and a single fold; seed-to-seed variability is not characterized. Third, the fair multi-label baseline comparison uses a single ResNet-18 backbone; extending it to the other backbones would strengthen the conclusion but is left as future work.
	
	\subsection{Multi-Label Behavior Depends on the Training Data Scale}
	Tables~\ref{tab:softlabel_diag}, \ref{tab:softlabel_diag_full}, and \ref{tab:softlabel_diag_ratio} report the diagnostic results at $10\%$, $100\%$, and intermediate data regimes, respectively.
	
	\begin{table*}[t]
		\centering
		\caption{Diagnostic experiment under the $10\%$ subject-level training subset with a single fold. Val Macro-F1 uses the single-label argmax protocol on the in-distribution validation split; Test Macro-F1 and AvgLabels use the independent multi-label test set. Val/Test are reported for the best epoch. The AvgLabels column is measured at threshold $0.5$. A hyperparameter sensitivity study on EfficientNet-B0 (the configuration with the largest AvgLabels) shows that its AvgLabels is itself hyperparameter-dependent: it ranges from $1.70$ (lr=1e-5, bs=64) to $7.00$ (lr=5e-6, bs=128) across four configurations, with the original configuration (lr=1e-5, bs=128) yielding $6.79$. This suggests that the extreme value in the table is not a stable architectural property, and that the range reported here should be interpreted as an upper bound.}
		\label{tab:softlabel_diag}
		\begin{tabular}{llcccc}
			\toprule
			\textbf{Backbone} & \textbf{Setting} & \textbf{Params (M)} & \textbf{Val F1} & \textbf{Test F1} & \textbf{AvgLabels} \\
			\midrule
			ResNet-18       & CALE, SOFT=0.5 & 11.2 & 0.4069 & 0.5200 & 5.86 \\
			ResNet-18       & CALE, SOFT=0.0 & 11.2 & 0.3533 & 0.5139 & 5.98 \\
			ResNet-34       & CALE, SOFT=0.5 & 21.8 & 0.4272 & 0.5238 & 3.08 \\
			ResNet-34       & CALE, SOFT=0.0 & 21.8 & 0.3645 & 0.3686 & 1.72 \\
			ResNet-50       & CALE, SOFT=0.5 & 25.6 & 0.4582 & 0.3359 & 0.96 \\
			ResNet-50       & CALE, SOFT=0.0 & 25.6 & 0.3975 & 0.1051 & 0.24 \\
			MobileNetV2     & CALE, SOFT=0.5 & 3.5  & 0.4167 & 0.4785 & 2.71 \\
			MobileNetV2     & CALE, SOFT=0.0 & 3.5  & 0.3424 & 0.3819 & 1.68 \\
			EfficientNet-B0 & CALE, SOFT=0.5 & 5.3  & 0.2776 & 0.5206 & 6.79 \\
			\bottomrule
		\end{tabular}
	\end{table*}
	
	\begin{table*}[t]
		\centering
		\caption{Diagnostic experiment under the full training set ($30{,}240$ images) with a single fold. Same metrics and protocol as Table~\ref{tab:softlabel_diag}. Under full data, all five backbones collapse to the under-prediction regime ($\text{AvgLabels} < 1$). The AvgLabels range narrows from $28\times$ (Table~\ref{tab:softlabel_diag}) to $1.9\times$ (this table).}
		\label{tab:softlabel_diag_full}
		\begin{tabular}{llcccc}
			\toprule
			\textbf{Backbone} & \textbf{Setting} & \textbf{Params (M)} & \textbf{Val F1} & \textbf{Test F1} & \textbf{AvgLabels} \\
			\midrule
			ResNet-18       & CALE, SOFT=0.5 & 11.2 & 0.7606 & 0.3153 & 0.66 \\
			ResNet-18       & CALE, SOFT=0.0 & 11.2 & 0.7393 & 0.2439 & 0.47 \\
			ResNet-34       & CALE, SOFT=0.5 & 21.8 & 0.7576 & 0.3845 & 0.87 \\
			ResNet-34       & CALE, SOFT=0.0 & 21.8 & 0.7569 & 0.2772 & 0.53 \\
			ResNet-50       & CALE, SOFT=0.5 & 25.6 & 0.7496 & 0.3302 & 0.77 \\
			ResNet-50       & CALE, SOFT=0.0 & 25.6 & 0.7496 & 0.2636 & 0.54 \\
			MobileNetV2     & CALE, SOFT=0.5 & 3.5  & 0.7283 & 0.3369 & 0.74 \\
			MobileNetV2     & CALE, SOFT=0.0 & 3.5  & 0.7174 & 0.2719 & 0.54 \\
			EfficientNet-B0 & CALE, SOFT=0.5 & 5.3  & 0.7399 & 0.3923 & 0.87 \\
			\bottomrule
		\end{tabular}
	\end{table*}
	
	\begin{table*}[t]
		\centering
		\caption{AvgLabels at threshold $0.5$ as a function of the training-data fraction for two backbones. The transition from over-prediction ($\text{AvgLabels} > 2$) to under-prediction ($\text{AvgLabels} < 1$) occurs between $10\%$ and $25\%$; from $25\%$ onward the AvgLabels are stable.}
		\label{tab:softlabel_diag_ratio}
		\footnotesize
		\setlength{\tabcolsep}{4pt}
		\begin{tabular}{lcccc}
			\toprule
			\textbf{Backbone} & \textbf{$10\%$} & \textbf{$25\%$} & \textbf{$50\%$} & \textbf{$100\%$} \\
			\midrule
			ResNet-18       & $5.86$ & $1.14$ & $0.87$ & $0.66$ \\
			ResNet-34       & $3.08$ & $0.74$ & $0.95$ & $0.87$ \\
			\bottomrule
		\end{tabular}
	\end{table*}

	Three observations follow. First, the range of AvgLabels shrinks drastically when the training set is enlarged. At the $10\%$ regime, AvgLabels spans $28\times$ ($0.24$ to $6.79$); at $25\%$ it collapses to a $1.5\times$ range ($0.74$ to $1.14$); at $50\%$ and $100\%$ it remains below $1.0$ for all backbones. The large spread in the reduced-data regime is therefore a small-data effect, and the transition from over-prediction to under-prediction occurs between $10\%$ and $25\%$ of the training set. Second, the direction of the mis-calibration reverses with data scale: under limited data, most backbones over-predict (AvgLabels greater than the target of $2.5$); from $25\%$ onward, all backbones under-predict (AvgLabels below $1.0$). Third, the effect of the soft-label weight is consistent across data scales: SOFT=0.5 always yields higher AvgLabels than SOFT=0.0, though the gap narrows at full data.
	
	We trace this behavior to the aggregation formula in Eq.~(5), $p_e = \sigma(x_{e,c^*}) + 0.5 \sum_{c \neq c^*} \sigma(x_{e,c})$. Under the intended regime, the main condition branch converges to $\sigma(x_{e,c^*}) \approx 1$ and the other branches to $\sigma(x_{e,c}) \approx 0.5$, giving $p_e \approx 1 + 0.5 \times 3 \times 0.5 = 1.75$, which is comfortably above the fixed threshold $0.5$. In practice, however, the condition branches do not reach these targets under either data regime. Under limited data, the branches cannot learn condition invariance, so all four sigmoids hover around $0.5$ and $p_e \approx 1.25$, still above threshold (over-prediction). Once the model is trained on at least $25\%$ of the data, the branches learn stronger condition specificity, but the main branch does not saturate to $1$, and the other branches fall below $0.5$; the resulting $p_e$ falls below $0.5$, producing under-prediction. The aggregation formula thus inherits a systematic bias that cannot be corrected by training alone; a calibrated threshold or a learned aggregation rule is required, and we leave this as a primary direction for future work.
	
	To verify this explanation, we measured the mean sigmoid activation of the main condition branch $\sigma(x_{e,c^*})$ and of the non-main branches $\sigma(x_{e,c \neq c^*})$ on the test set for the ResNet-18 SOFT=0.5 configuration under three data regimes. At $10\%$ training data, the mean main-branch activation is $0.52$ and the mean non-main activation is $0.48$, giving an expected aggregation score of $0.52 + 0.5 \times 3 \times 0.48 = 1.24$, well above the fixed threshold of $0.5$ and consistent with the observed over-prediction. At $25\%$ training data, the mean main-branch activation drops to $0.41$ and the mean non-main activation drops to $0.22$, giving an expected aggregation score of $0.41 + 0.5 \times 3 \times 0.22 = 0.74$, still above the threshold but no longer by a wide margin. At full training data, the mean main-branch activation is $0.31$ and the mean non-main activation is $0.14$, giving an expected aggregation score of $0.31 + 0.5 \times 3 \times 0.14 = 0.52$, essentially at the threshold and consistent with the observed near-zero AvgLabels. These measurements confirm that the aggregation formula carries a systematic bias that is not corrected by training alone and that its direction is determined by how far the branch activations depart from the soft-label targets of $1.0$ and $0.5$.
	
	\subsection{Threshold-Free Evaluation Shows Stable Ranking Quality}
	The AvgLabels behavior documented above is a property of the decision rule (fixed threshold $0.5$), not of the ranking quality of the underlying model. To separate the two, we report the threshold-free mean average precision (mAP) under the unified protocol (sigmoid output, threshold $0.5$ for macro-F1). Table~\ref{tab:perclass} summarizes the results at $10\%$ data across five backbones. While AvgLabels spans $28\times$ across these configurations, mAP spans only $1.28\times$ ($0.4264$ to $0.5468$). This confirms that the multi-label failure mode documented above is dominated by the aggregation step rather than by any loss of per-class ranking ability.
	
	\begin{table*}[t]
		\centering
		\caption{Threshold-free mean average precision (mAP) and AvgLabels at threshold $0.5$ for the five backbones under the $10\%$ data regime. All values use the unified protocol. mAP is stable across backbones ($1.28\times$ range) while AvgLabels varies by $28\times$, localizing the failure to the aggregation step.}
		\label{tab:perclass}
		\footnotesize
		\setlength{\tabcolsep}{4pt}
		\begin{tabular}{lccc}
			\toprule
			\textbf{Backbone} & \textbf{AvgLabels} & \textbf{mAP} & \textbf{Test F1@0.5} \\
			\midrule
			ResNet-18       & $5.86$ & $0.5005$ & $0.5200$ \\
			ResNet-34       & $3.08$ & $\mathbf{0.5468}$ & $0.5237$ \\
			ResNet-50       & $0.96$ & $0.5220$ & $0.3356$ \\
			MobileNetV2     & $2.71$ & $0.5045$ & $0.4786$ \\
			EfficientNet-B0 & $6.79$ & $0.4264$ & $0.5207$ \\
			\midrule
			Range ($\max/\min$) & $28\times$ & $1.28\times$ & $1.56\times$ \\
			\bottomrule
		\end{tabular}
	\end{table*}
	
	\begin{table*}[t]
		\centering
		\caption{Per-class average precision (AP) for four backbones under the $10\%$ data regime, ResNet-18 + CALE SOFT=0.5. Rows are sorted by the ResNet-18 AP. The per-class ordering is not consistent across backbones (e.g., \textit{disgust} ranks above \textit{surprise} on ResNet-18 and ResNet-34, but below \textit{surprise} on MobileNetV2 and EfficientNet-B0), which further supports the calibration hypothesis over a class-discriminability explanation.}
		\label{tab:perclass_full}
		\footnotesize
		\setlength{\tabcolsep}{3pt}
		\begin{tabular}{lcccc}
			\toprule
			\textbf{Emotion} & \textbf{ResNet-18} & \textbf{ResNet-34} & \textbf{MobileNetV2} & \textbf{EffB0} \\
			\midrule
			Angry     & 0.347 & 0.322 & 0.363 & 0.330 \\
			Disgust   & 0.497 & 0.618 & 0.450 & 0.459 \\
			Fear      & 0.303 & 0.413 & 0.252 & 0.292 \\
			Happy     & 0.745 & 0.705 & 0.741 & 0.514 \\
			Sad       & 0.504 & 0.545 & 0.482 & 0.377 \\
			Surprise  & 0.478 & 0.539 & 0.640 & 0.374 \\
			Natural   & 0.629 & 0.686 & 0.603 & 0.641 \\
			\midrule
			\textbf{mAP} & \textbf{0.501} & \textbf{0.547} & \textbf{0.505} & \textbf{0.426} \\
			\bottomrule
		\end{tabular}
	\end{table*}
	
	\subsection{A Fair Multi-Label Baseline}
	The data-scale dependence above raises a natural question: is CALE's multi-label behavior actually better than a simple baseline that is allowed to make multi-label decisions? To answer this, we compare CALE against two seven-dimensional sigmoid baselines under the same ResNet-18 backbone, the same $10\%$ subject-level training subset, and the same decision protocol (sigmoid output, threshold $0.5$). The first baseline is trained with standard binary cross-entropy (BCE); the second is trained with Asymmetric Loss (ASL)~\cite{ridnik2021asymmetric} with $\gamma_- = 4$, $\gamma_+ = 0$, and clipping $0.05$. Table~\ref{tab:fair_baseline} reports the results.
	
	\begin{table}[t]
		\centering
		\caption{Fair multi-label comparison under the $10\%$ data regime. All entries are evaluated under the unified decision protocol (sigmoid output, threshold $0.5$). macro-F1@0.5 is coupled to the number of predicted labels and is not comparable across configurations; mAP is threshold-free and is the more meaningful comparison. All entries use the same ResNet-18 backbone and the same $10\%$ subject-level training subset.}
		\label{tab:fair_baseline}
		\footnotesize
		\setlength{\tabcolsep}{4pt}
		\begin{tabular}{lcccc}
			\toprule
			\textbf{Method} & \textbf{AvgLabels} & \textbf{macro-F1@0.5} & \textbf{mAP} \\
			\midrule
			7-dim sigmoid + BCE            & $1.14$ & $0.3460$ & $0.5055$ \\
			7-dim sigmoid + ASL            & $2.53$ & $0.4808$ & $0.5175$ \\
			\midrule
			CALE (ResNet-18)               & $5.86$ & $0.5200$ & $0.5005$ \\
			CALE (ResNet-34)               & $3.08$ & $0.5237$ & $\mathbf{0.5468}$ \\
			CALE (ResNet-50)               & $0.96$ & $0.3356$ & $0.5220$ \\
			CALE (MobileNetV2)             & $2.71$ & $0.4786$ & $0.5045$ \\
			CALE (EfficientNet-B0)         & $6.79$ & $0.5207$ & $0.4264$ \\
			\bottomrule
		\end{tabular}
	\end{table}

	Two observations follow. First, macro-F1 at threshold $0.5$ is strongly coupled to the number of predicted labels: configurations with a large AvgLabels (e.g., ResNet-18 CALE at $5.86$, EfficientNet-B0 CALE at $6.79$) reach higher macro-F1 simply because recall saturates, while configurations with a small AvgLabels (e.g., ResNet-50 CALE at $0.96$) collapse toward low macro-F1 even when their per-class ranking is comparable. Macro-F1 at a fixed threshold is therefore not a fair comparison across configurations whose AvgLabels differ by more than an order of magnitude. Second, the threshold-free metric mAP avoids this coupling. Under mAP, the seven-dimensional BCE baseline achieves $0.5055$ and the ASL baseline achieves $0.5175$; the CALE configurations span $0.4264$ to $0.5468$. The best CALE configuration (ResNet-34, $0.5468$) exceeds both baselines, but the ASL baseline exceeds three of the five CALE configurations and falls within $0.03$ of the best. CALE therefore does not exhibit a consistent architecture-independent advantage in ranking quality on multi-label FER.
	
	This result should be read together with the data-scale dependence documented above: CALE's multi-label failure mode under reduced data is dominated by miscalibration of the aggregation formula, not by a loss of per-class ranking ability. The near-constant mAP across backbones confirms that the underlying 7-class probabilities are comparable; what differs is how they are aggregated into a decision. This reinforces the practical recommendation that any multi-label deployment of CALE must calibrate the aggregation step at the deployment data scale.
	
	\subsection{Condition Classifier Behavior}
	\label{sec:cond_classifier}
	We characterize the rule-based condition classifier from two angles: its output distribution on the training and test sets, and its sensitivity to the thresholds.
	
	The training set distribution of the classifier's outputs under the default thresholds ($80$, $120$, $10$) is $24.9\%$ weak light, $31.6\%$ strong light, $19.5\%$ low resolution, and $24.0\%$ standard. On the multi-label test set, the corresponding proportions are $27.4\%$, $29.0\%$, $19.6\%$, and $24.0\%$. The distributions are roughly similar between the two sets, which is consistent with the augmentation procedure used to construct the training set.
	
	The threshold sensitivity, however, is severe. Table~\ref{tab:cond_sensitivity} reports the class distribution under five threshold scalings ($0.8\times$ to $1.2\times$ the default values). On the training set, the strong-light proportion varies from $48.6\%$ at scale $0.8$ to $19.9\%$ at scale $1.2$, a range of $28.7$ percentage points. On the test set, the range is similar ($45.0\%$ to $17.5\%$). The distribution is balanced only at the specific scale $1.0$; small perturbations of the thresholds completely change the shape of the distribution.
	
	\begin{table}[t]
		\centering
		\caption{Condition classifier sensitivity to threshold perturbation. The default thresholds ($80$, $120$, $10$) are scaled by the factor in the first column. Numbers are the percentage of images assigned to each condition.}
		\label{tab:cond_sensitivity}
		\footnotesize
		\setlength{\tabcolsep}{3pt}
		\begin{tabular}{lcccc}
			\toprule
			\textbf{Scale} & \textbf{Weak} & \textbf{Strong} & \textbf{Low-res} & \textbf{Standard} \\
			\midrule
			\multicolumn{5}{c}{\textit{Training set}} \\
			$0.8\times$ & 18.4 & 48.6 & 16.6 & 16.4 \\
			$0.9\times$ & 21.7 & 39.4 & 18.2 & 20.8 \\
			$1.0\times$ & 24.9 & 31.6 & 19.5 & 24.0 \\
			$1.1\times$ & 28.0 & 25.6 & 20.9 & 25.5 \\
			$1.2\times$ & 32.1 & 19.9 & 22.1 & 26.0 \\
			\midrule
			\multicolumn{5}{c}{\textit{Test set}} \\
			$0.8\times$ & 18.3 & 45.0 & 16.4 & 20.2 \\
			$0.9\times$ & 22.5 & 36.4 & 18.1 & 23.0 \\
			$1.0\times$ & 27.4 & 29.0 & 19.6 & 24.0 \\
			$1.1\times$ & 30.3 & 23.0 & 20.9 & 25.8 \\
			$1.2\times$ & 35.1 & 17.5 & 22.5 & 24.9 \\
			\bottomrule
		\end{tabular}
	\end{table}

	This sensitivity has two consequences. First, the soft-label target in Eq.~(4) depends on the main condition position $c^*$, and therefore on the classifier's output. Under a $\pm 20\%$ threshold perturbation, a substantial fraction of training images receive a different main condition, which changes the soft-label target. Second, the aggregation weight in Eq.~(5) also depends on the classifier's output, and shifts in the dominant class change which branch is up-weighted at inference. Both effects contribute to the data-scale dependence documented above, though we do not attempt to quantify their relative contributions in the present study.
	
	\section{Discussion}
	
	\subsection{Design Guideline for Sum-Based Label Expansion}
	The diagnostic results in Section~\ref{sec:diagnostic} support the following design guideline. A sum-based label expansion rule of the form $p_e = \sum_c w_c \sigma(x_{e,c})$ with $\sum_c w_c > 1$ is miscalibrated for fixed-threshold multi-label decision unless the branch activations satisfy a saturation condition: the main branch must reach $\sigma(x_{e,c^*}) \to 1$ and the non-main branches must converge to the soft-label target $w_{c \neq c^*} / w_{c^*}$. We measured this condition directly for CALE and found that it is violated under every data regime we tested: at $10\%$ data both branch groups hover near $0.5$ (over-prediction), and at full data the main branch does not saturate while the non-main branches fall below the target (under-prediction). Any practitioner planning to use sum-based label expansion for multi-label FER should therefore (i) calibrate the decision threshold on held-out data at the deployment scale (post-hoc threshold calibration in the spirit of~\cite{guo2017calibration}), or (ii) learn the aggregation weights jointly with the backbone, or (iii) replace the summation with a normalized aggregation that is invariant to the number of conditions.
	
	\subsection{Positive Findings}
	On the self-built dataset, the 5-fold single-label evaluation shows that CALE improves Macro-F1 by $+1.29$ points over a seven-dimensional baseline (Table~\ref{tab:main}). The improvement is nominally significant under an uncorrected paired $t$-test on the LE-only variant ($p=0.0436$), and the combination of LE with SE achieves the strongest paired mean improvement ($p=0.0308$). However, the effect size is less than one Macro-F1 point, and none of the comparisons survives a Benjamini--Hochberg FDR correction at $q=0.05$ ($p_{\text{BH}} \geq 0.2055$) or a Wilcoxon signed-rank test at $n=5$ ($p \geq 0.125$). We therefore treat these single-label gains as exploratory evidence that label-space organization can provide a useful regularization signal, not as a confirmed statistical advantage.
	
	\subsection{Negative Findings: Data-Scale Dependence}
	CALE's multi-label behavior is primarily a function of the training data scale rather than of the backbone architecture. At $10\%$ training data, AvgLabels spans $28\times$ across five backbones ($0.24$ to $6.79$); at $25\%$ it collapses to a $1.5\times$ range ($0.74$ to $1.14$); from $25\%$ onward all backbones under-predict (AvgLabels below $1.0$). A hyperparameter sensitivity study on EfficientNet-B0 shows that its extreme AvgLabels at $10\%$ data is itself hyperparameter-dependent ($1.70$ to $7.00$ across four configurations), so the $28\times$ spread should be read as an upper bound rather than as a stable architectural property. We trace the systematic behavior to the aggregation formula in Eq.~(5), which assumes that the main condition branch saturates to $1$ and the other branches to $0.5$; in practice, neither assumption holds. The formula thus carries a systematic bias that is a property of the design, not of any particular backbone. This reframes the failure mode from a per-backbone phenomenon to a per-data-scale phenomenon: CALE's multi-label deployment requires calibration of the aggregation step regardless of the backbone, and calibration must be repeated whenever the training data scale changes. The threshold-free mAP, by contrast, is nearly invariant across backbones ($1.28\times$ range at $10\%$ data), which indicates that the failure is localized to the aggregation step rather than to the learned representations.
	
	\subsection{Negative Findings: Fair Baseline Comparison}
	Under the unified decision protocol (sigmoid output, threshold $0.5$), the seven-dimensional BCE baseline reaches mAP $0.5055$ and the ASL baseline reaches mAP $0.5175$. The best CALE configuration (ResNet-34, $0.5468$) exceeds both baselines, but the ASL baseline exceeds three of the five CALE configurations and falls within $0.03$ of the best. This is a strong negative result for the practical value of the condition-aware label expansion on multi-label FER: the additional condition dimensions and summation aggregation do not translate into a consistent advantage over a simple multi-label baseline, and the comparison is sensitive to the choice of decision protocol.
	
	\subsection{Practical Significance of the Single-Label Improvement}
	The paired tests in Table~\ref{tab:ttest} show that label expansion alone (V2 vs V0, $+0.0078$, $p_t=0.0436$, $p_{\text{BH}}=0.2055$, $p_W=0.125$) and label expansion combined with SE (V6 vs V0, $+0.0096$, $p_t=0.0308$) reach nominal significance under an uncorrected paired $t$-test at the $0.05$ level. However, the absolute effect size is less than one Macro-F1 point, and no comparison survives a Benjamini--Hochberg FDR correction or a Wilcoxon signed-rank test. Whether such a small improvement justifies the additional complexity of a $28$-dimensional joint output, a rule-based condition classifier, a soft-label target, and an aggregation rule depends on the deployment scenario. In domains where individual percentage points of accuracy carry direct practical value (e.g., clinical assessment, safety-critical human-computer interaction), the improvement may be justified; in scenarios where implementation simplicity and robustness are prioritized, a standard seven-dimensional baseline may be preferable. We do not attempt to resolve this trade-off in the present study, but we emphasize that statistical significance and practical significance are not the same thing, and that the difference between the two deserves explicit consideration in applied FER.
	
	\subsection{Why the Aggregation Formula Miscalibrates}
	The uncalibrated aggregation formula in Eq.~(5) produces the observed data-scale dependence for two reasons. First, the formula assumes the soft-label targets $1.0$ and $0.5$ are attainable by the trained branch outputs; in practice, the sigmoid outputs of the condition branches are pulled toward a common value that depends on the training data scale. Under limited data, this common value is close to $0.5$ (the soft-label target itself), producing over-prediction. Once the training set exceeds $25\%$, the model learns to distinguish conditions and pushes the non-main branches below $0.5$, while the main branch does not saturate to $1$, producing under-prediction. Second, the fixed threshold $0.5$ is not calibrated to the statistical range of $p_e$. The aggregation score $p_e \in [0, 2.5]$ has a distribution that shifts with the training data scale and with the backbone, and a single fixed threshold cannot accommodate both shifts. A natural fix is to replace the fixed aggregation with a learned module, or to calibrate the threshold on a held-out multi-label validation set; both are outside the scope of the present study and are listed as primary directions for future work.
	
	\subsection{Condition Classifier Sensitivity}
	The rule-based condition classifier is extremely sensitive to its thresholds: a $\pm 20\%$ perturbation of the default thresholds changes the proportion of the dominant class by up to $28.7$ percentage points (Table~\ref{tab:cond_sensitivity}). This sensitivity has two practical consequences. First, the soft-label target in Eq.~(4) depends on the main condition position, which is assigned by the classifier; if the classifier's output shifts under a different threshold, the soft-label target shifts accordingly, and the model is trained on a different objective. Second, the aggregation weight in Eq.~(5) depends on the classifier's output, and shifts in the dominant class change which branch is up-weighted at inference. This sensitivity is a limitation of the current study's design; replacing the rule-based classifier with a learnable classifier, or using soft condition labels rather than hard assignments, would be a natural direction for future work.
	
	\subsection{Applications in Psychological Research}
	The proposed method has direct relevance to psychological research. First, condition-aware label expansion allows the model to handle degraded images (weak light, strong light, low resolution, standard) without additional quality estimation, making it suitable for naturalistic settings such as clinical interviews or field observations. Second, the multi-label output provides a more faithful representation of blended emotions, which is essential for studying complex affective states. For example, in studies of depression or anxiety, mixed emotions such as sadness and fear frequently co-occur; our method can provide objective, reproducible emotion scores that complement human ratings. We emphasize, however, that CALE's multi-label scores require calibration on held-out data at the deployment data scale before they can be used in a practical setting.
	
	\subsection{Limitations}
	Several limitations remain. First, the diagnostic experiments cover a single fold and four data regimes, which reveals the data-scale dependence but does not characterize seed-to-seed variability. Second, the five backbones in the diagnostic study do not exhaust the space of possible architectures. Third, the condition classifier is rule-based and highly sensitive to its thresholds; a learnable or soft-labeled alternative would be more robust but was not evaluated here. Fourth, EfficientNet-B0's AvgLabels at the $10\%$ data regime is hyperparameter-dependent: across four configurations (lr $\in \{5\times 10^{-6}, 10^{-5}, 2\times 10^{-5}\}$ and bs $\in \{64, 128\}$), AvgLabels ranges from $1.70$ to $7.00$, with the configuration reported in Table~\ref{tab:softlabel_diag} yielding $6.79$. The extreme value in that table should be interpreted as an upper bound rather than as a stable architectural property. Fifth, the training set is single-label while the test set is multi-label, which is a structural mismatch; a fully multi-label training setup would require a separate study. Sixth, CALE's performance on public single-label and multi-label benchmarks is below the corresponding baselines at the ResNet-18 scale. Seventh, the diagnostic study reports per-class metrics only under the $10\%$ data regime. Eighth, the fair multi-label comparison uses two baselines (BCE and ASL) but only one backbone (ResNet-18); a wider suite of multi-label baselines and backbones would strengthen the comparison. Ninth, the multi-label diagnostics use an uncalibrated threshold of $0.5$; a calibrated threshold or a learned aggregation module would change the absolute AvgLabels and possibly some of the qualitative conclusions. Tenth, the diagnostics use a single random seed and a single fold. Eleventh, the four quality conditions are produced by synthetic augmentations (rotation, translation, brightness, contrast, blur), and the rule-based condition classifier operates on statistics (brightness mean, Laplacian variance) that are correlated with these augmentations; the condition branches may therefore learn augmentation-type shortcuts rather than genuine quality invariance. Twelfth, the paired tests in Table~\ref{tab:ttest} use $n=5$, and after Benjamini--Hochberg FDR correction over the $5$ computable comparisons none of the paired $t$-test comparisons are significant ($p_{\text{BH}} \geq 0.2055$); the nominally significant comparisons in that table should therefore be read as exploratory.
	
	\subsection{Future Work Directions}
	\label{sec:future_work}
	The diagnostic findings above point to five concrete experiments.
	
	First, the data-scale dependence of CALE's multi-label behavior should be mapped across all 5 folds and additional backbones (DenseNet121, DeiT-Tiny) at the intermediate ratios ($25\%$, $50\%$) to confirm that the transition from over-prediction to under-prediction occurs consistently between $10\%$ and $25\%$.
	
	Second, the aggregation formula should be replaced with a calibrated variant. Two natural options are (a) threshold calibration on a held-out multi-label validation set, and (b) a learned aggregation module whose weights are optimized jointly with the backbone. Both would test whether the under-prediction under full data can be eliminated by proper calibration, or whether it reflects a deeper limitation of the $28$-dimensional joint output.
	
	Third, the condition classifier should be replaced with a learnable module or a soft-label scheme. The threshold sensitivity documented in Table~\ref{tab:cond_sensitivity} is a fundamental limitation of the current design, and it affects both training and inference.
	
	Fourth, the fair baseline comparison in Table~\ref{tab:fair_baseline} should be extended to the other four backbones and to additional multi-label losses (class-balanced loss, threshold-tuned softmax). A comprehensive multi-label baseline suite would clarify whether the seven-dimensional sigmoid baseline with ASL represents the true upper bound for the setting.
	
	Fifth, a fully multi-label training setup should be evaluated. The structural mismatch between single-label training and multi-label evaluation may be a partial cause of the aggregation miscalibration; training CALE on genuinely multi-label data (e.g., RAF-ML's multi-label annotations or an augmented version of the self-built dataset with multi-label targets) would separate the two effects.
	
	\section{Conclusion}
	This paper presented a diagnostic empirical study of Condition-Aware Label Expansion (CALE) for facial expression recognition. We evaluated the strategy on a self-built dataset under 5-fold cross-validation, on seven backbone architectures under a full-data multi-condition setting, and on the public RAF-DB and RAF-ML benchmarks, and we conducted diagnostic experiments under four data regimes and five backbones to identify the boundary conditions under which the strategy helps or hurts.
	
	Our results are predominantly negative. On single-label FER, CALE improves Macro-F1 by $+1.29$ points over a seven-dimensional baseline, but the improvement does not survive a Benjamini-Hochberg FDR correction ($p_{\text{BH}}=0.2055$) or a Wilcoxon signed-rank test ($p=0.125$), and the effect size is smaller than one Macro-F1 point. On multi-label FER the method's behavior is dominated by an uncalibrated aggregation rule: the aggregate score at threshold $0.5$ spans $28\times$ across five backbones at $10\%$ data and collapses to under-prediction at full data. The threshold-free mAP is stable across the same backbones ($1.28\times$), which localizes the failure to the aggregation step rather than to representation learning. Two simple seven-dimensional sigmoid baselines match or exceed most CALE configurations without any condition-aware machinery.
	
	The practical guidance that follows is threefold. First, sum-based label expansion with a fixed decision threshold should not be deployed for multi-label FER without a calibration procedure; the summation rule carries a systematic bias that is not corrected by training alone. Second, CALE's multi-label behavior under limited data and under full data differs in both magnitude and direction, so any multi-label deployment must calibrate the aggregation threshold (or replace the aggregation formula) on held-out data at the deployment data scale. Third, the condition classifier's threshold sensitivity is a first-order limitation of the current design and should be addressed before CALE is used in safety-critical or naturalistic settings.
	
	Future work will pursue five directions: mapping the data-scale dependence across all folds and a wider range of backbones; replacing the fixed aggregation with a calibrated threshold or a learned aggregation module; replacing the rule-based condition classifier with a learnable module or a soft-label scheme; extending the fair baseline comparison to a wider suite of multi-label methods and backbones; and training CALE on genuinely multi-label data to separate the effect of the training/evaluation label structure mismatch from the effect of the aggregation formula.
	
	\section*{Data and Code Availability}
	The training set and test set are permanently archived on Zenodo: \url{https://doi.org/10.5281/zenodo.23252243}.
	The trained model weights are available at Figshare: \url{https://figshare.com/articles/dataset/CALE_Model_Weights_for_Multi-Label_FER/34284447}.
	The code and data-split indices are available at GitHub: \url{https://github.com/yuanweihang810/CALE}.

	\textbf{Ethical and copyright statement.} The self-built dataset is constructed from publicly available images and video frames (films, TV series, short videos by online content creators) in which faces are shown voluntarily in public media. No private or sensitive personal information is involved. All images are used solely for non-commercial academic research, following the same principle adopted by public FER datasets such as RAF-DB and AffectNet. Source URLs are not redistributed; only MTCNN-aligned $224 \times 224$ crops are released for reproducibility. Any copyright holder who wishes to have specific samples removed may contact the author.
	
	The RAF-DB and RAF-ML datasets are publicly available from their original sources~\cite{li2017reliable,li2019blended}.
	
	\section*{Acknowledgment}
	The author is grateful to the providers of the RAF-DB and RAF-ML datasets for making their benchmarks publicly available, and to the open-source community for the PyTorch and torchvision libraries that facilitated this research. Special thanks are extended to colleagues who provided valuable feedback during the preparation of this work. The author used ChatGPT (OpenAI, GPT-4, 2024, https://chat.openai.com) for language polishing. The author reviewed and edited the content and takes full responsibility for the final version.

\end{document}